# Load What You Need: Smaller Versions of Multilingual BERT

**Amine Abdaoui**
Geotrend / Toulouse
amine@geotrend.fr

**Camille Pradel**
Geotrend / Toulouse
camille@geotrend.fr

**Grégoire Sigel**
Geotrend / Toulouse
gregoire@geotrend.fr

## Abstract

Pre-trained Transformer-based models are achieving state-of-the-art results on a variety of Natural Language Processing data sets. However, the size of these models is often a drawback for their deployment in real production applications. In the case of multilingual models, most of the parameters are located in the embeddings layer. Therefore, reducing the vocabulary size should have an important impact on the total number of parameters. In this paper, we propose to generate smaller models that handle fewer number of languages according to the targeted corpora. We present an evaluation of smaller versions of multilingual BERT on the XNLI data set, but we believe that this method may be applied to other multilingual transformers. The obtained results confirm that we can generate smaller models that keep comparable results, while reducing up to 45% of the total number of parameters. We compared our models with DistilmBERT (a distilled version of multilingual BERT) and showed that unlike language reduction, distillation induced a 1.7% to 6% drop in the overall accuracy on the XNLI data set. The presented models and code are publicly available.

## 1 Introduction

While transformer based models are getting larger, there is a growing difficulty to meet production requirements when deploying them. Reducing the size of these models is therefore an important step towards a democratization of transformers in real industrial environments. In addition to the model architecture, the vocabulary size may have a huge impact on the total number of parameters. However, in the case of multilingual transformers, we need to increase the model vocabulary as we include more languages. For instance, the cased version of multilingual BERT (mBERT) has a vocabulary of 119k entries, while the english version (BERT$_{BASE}$) has only a 30k tokens vocabulary (Devlin et al., 2018). Therefore, even if both models share the same architecture, mBERT has 178 million parameters, while BERT$_{BASE}$ has only 110 million. As a matter of fact, mBERT allocates more than 51% of its parameters to the embeddings layer.

Still, reducing the vocabulary size may induce an important drop of the model performance on downstream tasks. Indeed, Conneau et al. (2020) obtained more than 3% improvement in overall accuracy on XNLI by increasing the vocab size from 128k to 512k. Here, we propose to reduce the vocabulary size by reducing the number of languages the model handles. Indeed, most of multilingual transformers have been learned on more than 100 languages. However, in several real world applications, we need to handle a lower number of languages. In this paper, we suggest to extract smaller multilingual transformers that handle fewer languages. Here, we evaluate smaller versions of mBERT on the XNLI data set (Conneau et al., 2018), but we believe that this method may be applied to other multilingual transformers and other NLP tasks.

We compared our models with the original mBERT and the hugging face multilingual DistilBERT (DistilmBERT) (Sanh et al., 2019). To our knowledge this is the first detailed evaluation of DistilmBERT on the XNLI data set. The obtained results confirm that unlike DistilmBERT, our strategy reduces the model size without decreasing the average accuracy. The aim of this work is to draw the attention of the community on this simple yet efficient way of reducing the size of multilingual transformers.

## 2 Related work

Several methods have recently emerged to compress transformer models.

A family of approaches focuses on quantizing model weights, i.e. reducing the memory footprint of a model by representing its weights by lower-precision values. This method, especially effective when used with specific hardware, has been recently applied by Shen et al. (2020) to the transformer architecture.

Knowledge distillation (Buciluǎ et al., 2006; Hinton et al., 2015) consists in transferring knowledge learned by a large teacher network to a smaller student. Works from this area aim at building models with simpler architectures than the original ones while mocking their behaviour. Knowledge distillation has been applied to reduce the number of layers of BERT models (Sun et al., 2019; Tang et al., 2019). Not limited to architecture simplification, Zhao et al. (2019) performs a model distillation by simultaneously training the teacher and student models in order to reduce the vocabulary size and the embeddings size.

Regarding multilingual transformers, Tsai et al. (2019) evaluated distilled versions of BERT and mBERT for POS tagging and Morphology tasks. Their version of mBERT is 6 times smaller and 27 times faster but induces an average F1 drop of 1.6% and 5.4% in the two evaluated tasks. Furthermore, the learnt model is not publicly available on the internet. In this paper, we compare our models with the widely used DistilmBERT (Sanh et al., 2019), a distilled version of mBERT that reduces its size by 21%.

Finally, fewer methods have tried to reduce the number of parameters located in the embeddings. Unlike Mehta et al. (2020), our method do not require to train the model from scratch.

## 3 Methods

In order to generate smaller versions of mBERT, we have (i) identified the vocabulary of each language, and then (ii) rebuilt the embedding layer to generate the corresponding models.

### 3.1 Selecting Language Vocabularies

As for the original mBERT, we started from the entire Wikipedia dump of each language[1]. In our case, we selected the 15 languages covered by the XNLI data set (Conneau et al., 2018). The recommended mBERT cased tokenizer has been used to tokenize the data. The frequency of each entry of the original mBERT vocabulary has been computed for each language. Manual evaluation of the tokens distributions over the different data sets allowed us to chose an appropriate frequency threshold. For each language, tokens appearing in at least 0.05% of its paragraphs (lines) were selected in its vocabulary. Table 1 presents the number of selected tokens for each language and their proportions in the original mBERT vocabulary. As expected, the number of selected tokens for the 15 languages (union) is not equal to the sum of selected tokens in each language. Indeed, several languages should share a certain number of tokens (proper nouns, punctuation signs, numbers, etc.).

[1] https://lindat.mff.cuni.cz/repository/xmlui/handle/11234/1-2735

| Language | #Selected tokens | Proportion of original tokens |
|---|---|---|
| English (en) | 28458 | 23.8% |
| French (fr) | 24482 | 20.5% |
| Spanish (es) | 26346 | 22.0% |
| German (de) | 26031 | 21.8% |
| Greek (el) | 11616 | 9.7% |
| Bulgarian (bg) | 12121 | 10.1% |
| Russian (ru) | 14270 | 11.9% |
| Turkish (tr) | 19086 | 16.0% |
| Arabic (ar) | 7292 | 6.1% |
| Vietnamese (vi) | 17512 | 14.6% |
| Thai (th) | 8493 | 7.1% |
| Chinese (zh) | 12928 | 10.8% |
| Hindi (hi) | 5664 | 4.7% |
| Swahili (sw) | 16619 | 13.9% |
| Urdu (ur) | 8656 | 7.2% |
| **Union (15 langs)** | **71577** | **60%** |

Table 1: The number of selected tokens for each language and the proportion it covers in the original mBERT vocabulary.

### 3.2 Generating Smaller Models

Once the tokens selected for each targeted language, we extracted their corresponding embeddings to generate smaller models. Except the embeddings selection and re-arranging, no other modification has been applied to the model parameters. We generated 30 models covering the 15 XNLI languages in different ways:

- One multilingual model covering all the 15 languages (mBERT$_{15langs}$).

- 14 bilingual models combining english with another language from the remaining ones (mBERT$_{en-xx}$);

- 15 monolingual models (mBERT$_{xx}$);

All these models have been uploaded to the transformers hub to facilitate their use by the NLP community[2]. They can be easily fine-tuned on downstream tasks as conducted in the following section. The data and code are also available on github to allow users to generate other configurations of multilingual transformers[3].

# 4 Results and Discussion

In this section, we present the results of the original mBERT, its available distilled version (DistilmBERT) and our extracted mBERT versions on the XNLI data set. We also discuss the obtained results and show a few limitations of the proposed method.

## 4.1 Results

The above cited models have been evaluated for Cross-lingual Natural Language Inference. We used the XNLI data set which is an extension of the MultiNLI corpus (Williams et al., 2018) to 15 languages. The original english development and test sets have been manually translated to the remaining 14 languages. Furthermore, the XNLI data set comes with other configurations where the items have been automatically translated. In this paper, we used all the proposed configurations:

- *Cross-lingual Transfer*: Cross-lingual transfer from an english training set;

- *Translate Train*: Translate the english training set in order to learn the models on the same language of the test data;

- *Translate Train-all*: Translate the english training set and learn the models on all languages;

- *Translate Test*: Translate each test set to english and learn on the english training data.

Table 2 presents the obtained accuracies as well as the number of parameters of each model. All the results presented here may be reproduced using the shared models and the evaluation scripts available in the *transformers* library (Wolf et al., 2019).

[2] https://huggingface.co/Geotrend
[3] https://github.com/Geotrend-research/smaller-transformers

## 4.2 Discussion

Overall, our extracted versions of mBERT give similar results to those of the original model while being between 21% and 45% smaller in size. Regarding disilmBERT, the results show an average drop of 1.7% to 6.1% in terms of accuracy while being 25% smaller than the original mBERT. The drop in DistilmBERT's accuracy is much higher in the case of *Cross-lingual Transfer* from english to other languages (6.1%). On the contrary, our extracted versions seem to be resilient over all configurations.

The average accuracy of mBERT$_{15langs}$ is always very close to the one obtained by the original mBERT except for the *Translate Train-all* configuration. In this case, the accuracy of mBERT$_{15langs}$ is higher by 1.1%. Conneau et al. (2020) reported similar observations showing that the average accuracy decreases when the number of languages goes from 15 to 100. However, the authors kept the same vocabulary size in both experiments (150k tokens). Therefore, the per-language vocabulary should be lower in the 100-languages model than in the one handling only 15 languages. In our experiments, the vocabulary size of mBERT$_{15langs}$ is 40% smaller than the one of mBERT since we were trying to select tokens that are frequent only in these languages. Another important difference is that we are starting from models that were already trained on more than 100 languages and just fine-tuning them on less languages. Here the obtained results may suggest that starting from a certain point, keeping tokens that are very rare (or non-existent) in some languages may harm the fine-tuning of multilingual transformers on these specific languages even if we increase the vocabulary capacity.

Bilingual models (mBERT$_{en-xx}$) have been evaluated for *Cross-lingual Transfer* from the original english training set to the human translated test sets. They obtained a similar average accuracy to the one obtained by mBERT. All the presented models show better accuracies when evaluated on languages that are somewhat similar to english such as: french (fr), spanish (es) and german (de).

Finally, monolingual models (mBERT$_{xx}$) have been evaluated when the training and test sets are in the same language (*Translate Train* and *Translate Test*). Their average accuracies are lower but very close to ones obtained by mBERT. The difference is so small (less than 0.2%) that we avoid making interpretations here.

| Models | #params | en | fr | es | de | el | bg | ru | tr | ar | vi | th | zh | hi | sw | ur | avg |
|---|---|---|---|---|---|---|---|---|---|---|---|---|---|---|---|---|---|
| *Fine-tune multilingual model on English training set (Cross-lingual Transfer)* | | | | | | | | | | | | | | | | | |
| mBERT | 178 M | 82.1 | 73.8 | 73.9 | 70.3 | **66.7** | 67.8 | 68.4 | 60.4 | 64.5 | 70.7 | 53.0 | 68.2 | **59.5** | 50.3 | 57.0 | 65.8 |
| DistilmBERT | 135 M | 78.5 | 70.2 | 70.1 | 65.3 | 60.4 | 63.1 | 63.9 | 55.8 | 58.6 | 57.4 | 37.2 | 64.2 | 51.6 | 46.7 | 53.3 | 59.7 |
| mBERT$_{15langs}$ | 141 M | **82.2** | **74.1** | 73.7 | 70.2 | 66.3 | **68.1** | 68.7 | **61.1** | 64.9 | 70.6 | 53.1 | **68.8** | 58.9 | 49.0 | 57.1 | 65.8 |
| mBERT$_{en-xx}$ | 108-113 M | **82.2** | 73.8 | **75.0** | **71.6** | 65.3 | **68.1** | **69.1** | 60.1 | **65.0** | **70.9** | **53.2** | 68.2 | **59.5** | 49.1 | **57.9** | **65.9** |
| *Fine-tune multilingual model on each training set (TRANSLATE-TRAIN)* | | | | | | | | | | | | | | | | | |
| mBERT | 178 M | 82.1 | 77.2 | 78.1 | **77.1** | 74.5 | 75.2 | **74.2** | 71.5 | 70.6 | **75.3** | 64.9 | 76.2 | 66.5 | **66.8** | 61.6 | 72.8 |
| DistilmBERT | 135 M | 78.5 | 74.6 | 75.6 | 72.9 | 71.0 | 70.9 | 70.4 | 68.1 | 66.0 | 71.1 | 60.5 | 72.7 | 63.1 | 62.0 | 60.3 | 69.2 |
| mBERT$_{15langs}$ | 141 M | **82.2** | **77.8** | **78.5** | 76.5 | 74.1 | **76.1** | 73.6 | **71.8** | **71.5** | 75.2 | **65.2** | 75.6 | **67.1** | 65.7 | **62.4** | **72.9** |
| mBERT$_{xx}$ | 90-108 M | 81.6 | **77.8** | 77.1 | 76.0 | 74.3 | 75.0 | **74.2** | 71.4 | 71.4 | 75.1 | 64.9 | **77.0** | 67.0 | 64.3 | 61.6 | 72.6 |
| *Fine-tune multilingual models on all training sets (TRANSLATE-TRAIN-ALL)* | | | | | | | | | | | | | | | | | |
| mBERT | 178 M | 81.3 | 77.5 | 78.0 | 75.6 | 75.0 | 75.7 | 74.2 | **72.9** | 72.0 | 75.5 | 66.4 | 76.3 | 68.9 | 66.7 | 65.2 | 73.4 |
| DistilmBERT | 135 M | 79.9 | 75.6 | 76.6 | 75.0 | 73.2 | 74.1 | 72.4 | 70.8 | 70.3 | 73.7 | 62.7 | 74.5 | 65.6 | 66.5 | 64.1 | 71.7 |
| mBERT$_{15langs}$ | 141 M | **82.7** | **78.2** | **79.1** | **77.8** | **76.1** | **77.6** | **75.5** | **72.9** | **72.9** | **76.4** | **66.9** | **77.9** | **70.2** | **67.1** | **66.1** | **74.5** |
| *Translate everything to English and use English-only model (TRANSLATE-TEST) )* | | | | | | | | | | | | | | | | | |
| mBERT | 178 M | 82.1 | 74.9 | **76.6** | 73.7 | 74.2 | **76.7** | **71.3** | 70.8 | 70.4 | 68.4 | **66.6** | **70.9** | 65.8 | 61.9 | **63.0** | **71.2** |
| DistilmBERT | 135 M | 78.5 | 72.9 | 73.9 | 72.4 | 72.3 | 73.4 | 69.4 | 69.1 | 69.1 | 67.5 | 65.7 | 69.0 | 65.1 | 61.7 | 62.2 | 69.5 |
| mBERT$_{15langs}$ | 141 M | **82.2** | **75.3** | 76.3 | **74.5** | **74.6** | 75.8 | 70.9 | 70.6 | **71.3** | 68.6 | **66.6** | 70.8 | **65.8** | 62.0 | 61.9 | 71.1 |
| mBERT$_{en}$ | 108 M | **82.2** | 74.4 | 76.4 | 73.8 | 74.2 | 75.9 | 71.1 | **71.1** | 70.5 | **69.1** | 65.9 | 70.2 | 65.7 | **62.1** | 62.3 | 71.0 |

Table 2: Results on the XNLI data set of mBERT, its distilled version (DistilmBERT) and our extracted smaller versions mBERT$_{15langs}$, mBERT$_{en-xx}$ and mBERT$_{xx}$. We also report the number of parameters of each model.

## 4.3 Limitations

In this work, we were interested in reducing the number of parameters of multilingual transformer models, which leads to smaller models that require less memory. We believe that memory limits are crucial especially when deploying transformers on public cloud platforms. Moreover, smaller models are loaded faster than larger ones which may also improve the speed of deployed applications. However, the proposed method does not improve the inference speed since the model architecture has not changed. Whereas distillation allows to build smaller models that usually run faster. For example, DistilmBERT reduces the number of layers by a factor of 2 (from 12 to 6 layers), which also reduces the number of operations executed either during training or inference.

Table 3 presents the model size, the allocated memory, the loading time and the inference time for all the evaluated versions of mBERT. All these measurements have been computed on a Google Cloud *n1-standard-1*[4] machine (1 vCPU, 3.75 GB). As expected, our extracted models allow to reduce the first three measurements but without changing the inference time. Whereas ditilmBERT, in addition to reducing the size, the memory and the loading time, also enhances the inference speed by a factor of 2. Our experiments confirm that reduc-ing the number of embeddings and therefore the cost of the lookup operation has almost no impact on the inference time.

---

[4] https://cloud.google.com/compute/docs/machine-types#n1_machine_type

| Model | Size (MB) | Memory (MB) | Loading (sec) | Inference (sec) |
|---|---|---|---|---|
| mBERT | 714 | 1401 | 4.18 | 0.24 |
| DistilmBERT | 542 | 1070 | 3.08 | **0.12** |
| mBERT$_{15langs}$ | 564 | 1098 | 3.14 | 0.24 |
| mBERT$_{en\_xx}$ | 445 | 860 | 2.76 | 0.24 |
| mBERT$_{xx}$ | **393** | **760** | **2.45** | 0.24 |

Table 3: The model size, the allocated memory, the loading time and the inference time for all the evaluated versions of mBERT. We present the average measurements for bilingual and monolingual models. The loading times were computed 10 times for each model, while the inference times were averaged over 100 items from the XNLI data set (batch size = 1).

That being said, we can still apply language reduction to distilled transformers to take advantage of both methods and make our models even smaller.

## 5 Conclusion

Multilingual transformers have several advantages such as their capacity to do zero shot cross-lingual transfer. Therefore, most of multilingual transformers have been learnt to handle an important number of languages (around 100 languages). However, handling more languages requires to increase the vocabulary capacity and therefore the model size.

In this paper, we evaluated a simple method to break multilingual transformers into smaller models according to the targeted languages. We evaluated smaller versions of mBERT on the XNLI data set and showed that they reduced the number of parameters without decreasing the average accuracy.

As a future work, it would be interesting to evaluate this method on more models and tasks. Indeed, we are planning to reduce more recent multilingual transformers that showed better results than mBERT such as XLM-R (Conneau et al., 2020). We hope that these models will facilitate the deployment of multilingual transformers in real world applications.

## References

Cristian Buciluă, Rich Caruana, and Alexandru Niculescu-Mizil. 2006. Model compression. In *Proceedings of the 12th ACM SIGKDD international conference on Knowledge discovery and data mining*, pages 535–541.

Alexis Conneau, Kartikay Khandelwal, Naman Goyal, Vishrav Chaudhary, Guillaume Wenzek, Francisco Guzmán, Edouard Grave, Myle Ott, Luke Zettlemoyer, and Veselin Stoyanov. 2020. Unsupervised cross-lingual representation learning at scale. In *Proceedings of the 58th Annual Meeting of the Association for Computational Linguistics*, page 8440–8451.

Alexis Conneau, Ruty Rinott, Guillaume Lample, Adina Williams, Samuel R. Bowman, Holger Schwenk, and Veselin Stoyanov. 2018. Xnli: Evaluating cross-lingual sentence representations. In *Proceedings of the 2018 Conference on Empirical Methods in Natural Language Processing*. Association for Computational Linguistics.

Jacob Devlin, Ming-Wei Chang, Kenton Lee, and Kristina Toutanova. 2018. Bert: Pre-training of deep bidirectional transformers for language understanding. *arXiv preprint arXiv:1810.04805*.

Geoffrey Hinton, Oriol Vinyals, and Jeff Dean. 2015. Distilling the knowledge in a neural network. *arXiv preprint arXiv:1503.02531*.

Sachin Mehta, Rik Koncel-Kedziorski, Mohammad Rastegari, and Hannaneh Hajishirzi. 2020. Define: Deep factorized input token embeddings for neural sequence modeling. In *Proceedings of the 2020 International Conference on Learning Representations*.

Victor Sanh, Lysandre Debut, Julien Chaumond, and Thomas Wolf. 2019. Distilbert, a distilled version of bert: smaller, faster, cheaper and lighter. *arXiv preprint arXiv:1910.01108*.

Sheng Shen, Zhen Dong, Jiayu Ye, Linjian Ma, Zhewei Yao, Amir Gholami, Michael W Mahoney, and Kurt Keutzer. 2020. Q-bert: Hessian based ultra low precision quantization of bert. In *AAAI*, pages 8815–8821.

Siqi Sun, Yu Cheng, Zhe Gan, and Jingjing Liu. 2019. Patient knowledge distillation for bert model compression. *arXiv preprint arXiv:1908.09355*.

Raphael Tang, Yao Lu, Linqing Liu, Lili Mou, Olga Vechtomova, and Jimmy Lin. 2019. Distilling task-specific knowledge from bert into simple neural networks. *arXiv preprint arXiv:1903.12136*.

Henry Tsai, Jason Riesa, Melvin Johnson, Naveen Arivazhagan, Xin Li, and Amelia Archer. 2019. Small and practical bert models for sequence labeling. *arXiv preprint arXiv:1909.00100*.

Adina Williams, Nikita Nangia, and Samuel Bowman. 2018. A broad-coverage challenge corpus for sentence understanding through inference. In *Proceedings of the 2018 Conference of the North American Chapter of the Association for Computational Linguistics: Human Language Technologies, Volume 1 (Long Papers)*, pages 1112–1122. Association for Computational Linguistics.

Thomas Wolf, Lysandre Debut, Victor Sanh, Julien Chaumond, Clement Delangue, Anthony Moi, Pierric Cistac, Tim Rault, Rémi Louf, Morgan Funtowicz, Joe Davison, Sam Shleifer, Patrick von Platen, Clara Ma, Yacine Jernite, Julien Plu, Canwen Xu, Teven Le Scao, Sylvain Gugger, Mariama Drame, Quentin Lhoest, and Alexander M. Rush. 2019. Huggingface's transformers: State-of-the-art natural language processing. *ArXiv*, abs/1910.03771.

Sanqiang Zhao, Raghav Gupta, Yang Song, and Denny Zhou. 2019. Extreme language model compression with optimal subwords and shared projections. *arXiv preprint arXiv:1909.11687*.